\documentclass[letterpaper]{article} 
\usepackage{aaai2026}  
\usepackage{times}  
\usepackage{helvet}  
\usepackage{courier}  
\usepackage[hyphens]{url}  
\usepackage{graphicx} 
\usepackage{natbib}  
\usepackage{caption} 
\usepackage{algorithm}
\usepackage{algorithmic}
\usepackage{amsmath}
\usepackage{newfloat}
\usepackage{listings}
\DeclareCaptionStyle{ruled}{labelfont=normalfont,labelsep=colon,strut=off} 
\floatstyle{ruled}
\newfloat{listing}{tb}{lst}{}
\floatname{listing}{Listing}
\title{NeSTR: A Neuro-Symbolic Abductive Framework for Temporal Reasoning in Large Language Models}
\author{
    Feng Liang\textsuperscript{\rm 1}\thanks{The work was done during the first author's internship at National Key Laboratory of Big Data and Decision, National University of Defense Technology, China.}, 
    Weixin Zeng\textsuperscript{\rm 2}\thanks{Corresponding authors.}, 
    Runhao Zhao\textsuperscript{\rm 2}, 
    Xiang Zhao\textsuperscript{\rm 2}\footnotemark[2]
}

\affiliations{
    \textsuperscript{\rm 1}China Academy of Launch Vehicle Technology,\\

    \textsuperscript{\rm 2}National Key Laboratory of Big Data and Decision, National University of Defense Technology, China\\
    {fungloeng@gmail.com}, {\{zengweixin13, runhaozhao, xiangzhao\}@nudt.edu.cn}
}

\usepackage{bibentry}

\begin{document}

\maketitle

\begin{abstract}
Large Language Models (LLMs) have demonstrated remarkable performance across a wide range of natural language processing tasks. However, temporal reasoning, particularly under complex temporal constraints, remains a major challenge. To this end, existing approaches have explored symbolic methods, which encode temporal structure explicitly, and reflective mechanisms, which revise reasoning errors through multi-step inference. Nonetheless, symbolic approaches often underutilize the reasoning capabilities of LLMs, while reflective methods typically lack structured temporal representations, which can result in inconsistent or hallucinated reasoning. As a result, even when the correct temporal context is available, LLMs may still misinterpret or misapply time-related information, leading to incomplete or inaccurate answers. To address these limitations, in this work, we propose Neuro-Symbolic Temporal Reasoning (NeSTR), a novel framework that integrates structured symbolic representations with hybrid reflective reasoning to enhance the temporal sensitivity of LLM inference. NeSTR preserves explicit temporal relations through symbolic encoding, enforces logical consistency via verification, and corrects flawed inferences using abductive reflection. Extensive experiments on diverse temporal question answering benchmarks demonstrate that NeSTR achieves superior zero-shot performance and consistently improves temporal reasoning without any fine-tuning, showcasing the advantage of neuro-symbolic integration in enhancing temporal understanding in large language models.
\end{abstract}

\begin{links}
    \link{Code and Extended version}{https://github.com/fungloeng/NeSTR.git}
\end{links}

\section{Introduction}

Large Language Models (LLMs), such as GPT-4~\cite{achiam2023gpt}, Gemini2.5~\cite{comanici2025gemini}, and Qwen3~\cite{yang2025qwen3}, have demonstrated remarkable emergent capabilities, achieving human-level performance across a wide range of Natural Language Processing (NLP) tasks~\cite{zhao2023survey,salemi2023lamp}. The success of these models is rooted in pretraining on vast static corpora rich in world knowledge~\cite{roberts2020much}. However, the static nature of LLMs restricts their ability to answer time-sensitive queries, resulting in outdated or hallucinated responses when recent or temporally grounded information is required ~\cite{wang2024knowledge}. This limitation is particularly evident in Temporal Question Answering (TQA), which requires both \emph{timeliness}, i.e.,  accessing up-to-date knowledge, and \emph{temporal reasoning ability}, i.e., the ability to understand and use time expressions in context~\cite{wang2023tram,yang2024enhancing,zhao2025towards}. 

To satisfy the timeliness requirement of TQA, the retrieval-augmented generation (RAG) technique is utilized, which allows LLMs to access up-to-date external information at inference time~\cite{zhu2023large,chen2024benchmarking}. 
However, most existing approaches mainly focus on optimizing the retrieval pipeline, such as improving retrievers or re-rankers~\cite{wu2024time,qian2024timer4,chen2024bge,zhang2025qwen3}, while overlooking the importance of temporal reasoning~\cite{gupta2023temptabqa,wang2024knowledge}. 
As a result, models often fail to generate correct answers even when relevant evidence is available. 
Thus, it is critical to exploit the temporal information in the context to make accurate temporal reasoning ~\cite{jia2024faithful,su2024timo}.

To improve temporal reasoning ability, recent works mainly exploit two broad directions: \emph{symbolic structuring} and \emph{reflective reasoning}. 
The former transforms temporal information into structured forms, supporting explicit rule-based reasoning. For instance, QAaP parses questions and retrieved passages as Python-style dictionaries and applies programmable functions to verify consistency and rank answer candidates~\cite{zhu2023question}. 
Event-AL constructs event graphs from symbolic tuples and performs abductive reasoning over inferred temporal relations~\cite{wu2024event}. Reflective approaches, on the other hand, leverage the reasoning flexibility of LLMs by prompting them to reflect on intermediate steps. TISER, for example, guides the model through timeline construction and iterative revision, improving consistency on complex or unfamiliar temporal tasks~\cite{bazaga2025learning}.

However, symbolic methods, while accurate, are often fragile and heavily depend on predefined static logic templates, which makes it difficult for them to handle flexible time expressions and reasoning in natural language. 
In comparison, reflective methods offer greater flexibility, but they lack structural guidance, often resulting in reasoning errors and incomplete or inaccurate answers. 

To fill in this gap, in this work, we propose to integrate symbolic and reflective inference to balance interpretability and robustness—both crucial for handling diverse temporal questions.
Specifically, we propose NeSTR, a neuro-symbolic temporal reasoning prompting strategy that combines structured symbolic representations with neural inference. Temporal facts, such as events, relations, and time intervals, are first encoded as logical predicates, forming a transparent and interpretable basis for reasoning. 
Then, an LLM is employed as a neural reasoning engine, operating not over raw contexts but directly over the aforementioned temporal symbols.
Instead of using static rule-based systems, the model performs flexible multi-step inference by drawing on its internal knowledge to recognize temporal patterns, conducts abductive reasoning, and revises conclusions when inconsistencies are detected. 
This design enables NeSTR to integrate the precision and clarity of symbolic logic with the adaptability and generalization capabilities of neural models, thereby supporting accurate and explainable temporal question answering under complex constraints.

We evaluate NeSTR across multiple temporal QA benchmarks under zero-shot settings, demonstrating state-of-the-art performance without any model fine-tuning. Our experiments reveal that symbolic representations significantly enhance temporal reasoning, and abductive reflection effectively corrects logical inconsistencies, resulting in more robust and interpretable answers. 
The main contributions are:
\begin{itemize}
\item We propose \textbf{NeSTR}, the first neuro-symbolic prompting strategy for temporal question answering, integrating symbolic temporal representations with neural inference in large language models.
\item We design an interactive reasoning strategy in which symbolic feedback guides the neural inference process, enabling iterative error correction and step-by-step abductive reasoning within a unified neuro-symbolic prompting framework.
\item {We constrain the neural reasoning space of the language model by structured symbolic representation, ensuring the integrity of temporal information during inference.}
\end{itemize}

\section{NeSTR}

\begin{figure*}[t]
\centering
\includegraphics[width=\textwidth]{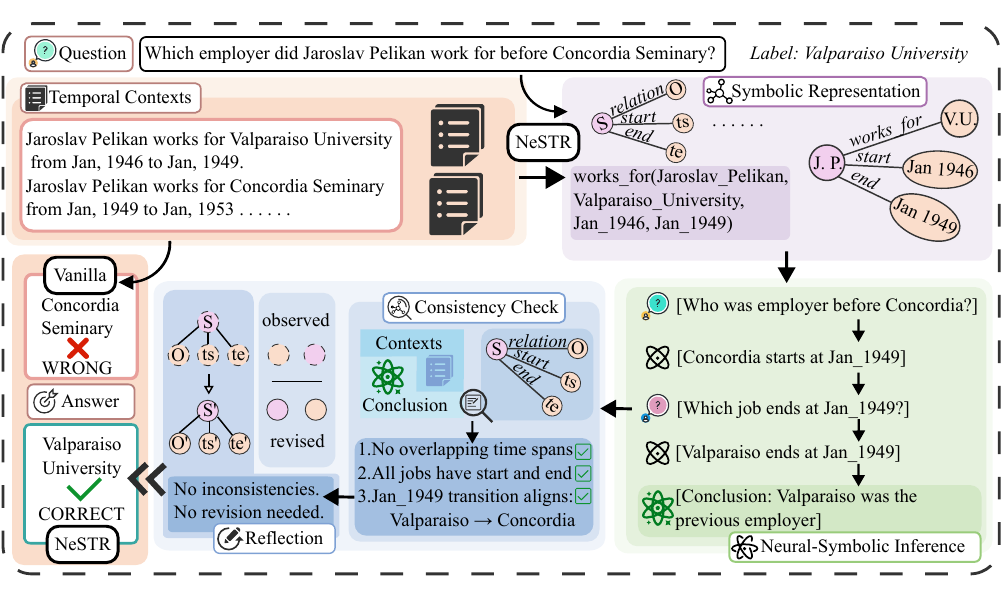}
\caption{Overview of NeSTR compared to vanilla inference on a temporal question. Given a question and temporal contexts (left), vanilla models often fail due to inadequate temporal reasoning. In contrast, NeSTR explicitly decomposes the reasoning process into symbolic representation, inference, consistency checking, and reflection, achieving accurate and temporally consistent answers.}
\label{fig:nestr_method}
\end{figure*}

\subsection{Problem Setting}

Formally, given a question $q$ associated with a specific temporal interval $[t_s, t_e]$ and a temporal context $c$ consisting of factual statements ${f_1, f_2, \dots, f_m}$ with explicit time stamps, the TQA task is to generate an answer $a$ that is both factually correct and temporally valid concerning the interval specified by $q$.
This task introduces two fundamental challenges: (1) accurately extracting and interpreting temporal expressions in natural language, and (2) reasoning logically over temporal facts, resolving conflicts, and generating answers consistent with the temporal constraints of the query. For instance, answering the question \textit{"Who did Jessica Valenti work for from 2005 to 2008?"} requires identifying all organizational affiliations overlapping with the specified period and discarding irrelevant temporal facts. Our work specifically targets the zero-shot setting, requiring the model to perform inference based solely on the provided context without any fine-tuning.

\subsection{NeSTR Architecture}

To address the challenge of interpretable and accurate reasoning under complex temporal constraints, we propose NeSTR, a neuro-symbolic prompting strategy that combines structured symbolic representations with neural inference via large language models. NeSTR first converts natural language inputs into logical predicates, forming an explicit and interpretable substrate for reasoning. A language model then performs flexible multi-step inference directly over these symbolic structures, enabling consistency checking, pattern recognition, and reflective correction. This architecture unifies the clarity of symbolic logic with the adaptability and contextual depth of neural models.

As shown in Figure~\ref{fig:nestr_method}, NeSTR follows a five-stage neuro-symbolic reasoning process: (1) Symbolic Representation: extract temporal facts and transform them as structured symbolic predicates; (2) Neural-Symbolic Inference: use temporal symbols as feedback to guide neural reasoning through iterative and abductive steps; (3) Consistency Verification: check logical and temporal coherence; (4) Reflection: revise intermediate results if inconsistencies are found; (5) Answer: produce a concise, evidence-based answer. This modular prompting strategy enhances both robustness and interpretability in temporal reasoning. We now describe each stage of the NeSTR prompting strategy in detail:

\paragraph{Symbolic Representation}
Temporal information in natural language is often vague, implicit, or expressed in diverse surface forms. This makes it difficult for language models to reason accurately about time. NeSTR addresses this issue by converting raw temporal context into structured symbolic representations. Each symbol captures an event or relation using a normalized format, such as {relation( subject, object, start\_time, end\_time )}. These representations are generated by extracting event entities and aligning them with explicit or implicit time expressions in the input. 

For example, a sentence like \textit{''From 1946 to 1949, Jaroslav Pelikan worked at Valparaiso University''} would be converted into {works\_for(Jaroslav\allowbreak Pelikan, Valparaiso\allowbreak University, 1946, 1949)}, which encodes the person, organization, and duration in a structured and machine-interpretable form. This symbolic representation preserves temporal boundaries in a normalized format and allows the model to reason about time in a consistent and logic-compatible manner.

Compared to prior reflective approaches that rely solely on reasoning over natural language, this symbolic representation step makes the temporal structure visible and tractable. The structured format reduces ambiguity and noise in the input while preserving complete temporal and semantic information, ensuring that key entities and time expressions are retained as a stable foundation for symbolic reasoning. Formally, each context fact is represented as a symbolic quadruple:

\begin{equation}
f_i = \textit{relation}(s_i, o_i, t^{(i)}_s, t^{(i)}_e)
\end{equation}

where $s_i$ and $o_i$ denote the subject and object entities, and $t^{(i)}_s, t^{(i)}_e$ are normalized numeric timestamps representing the start and end time of the fact. The entire context is thus converted into a symbolic fact set:

\begin{equation}
\mathcal{F} = \{ f_1, f_2, \dots, f_m \}
\end{equation}

Given a temporal question $q$ associated with a target interval $[t^{q}_{s}, t^{q}_{e}]$, NeSTR aims to identify the subset of facts $\mathcal{F}_q \subseteq \mathcal{F}$ that are temporally relevant:

\begin{equation}
\mathcal{F}_q = \{ f_i \in \mathcal{F} \mid [t^{(i)}_{s}, t^{(i)}_{e}] \cap [t^{q}_{s}, t^{q}_{e}] \neq \emptyset \}
\end{equation}

\paragraph{Neural-Symbolic Inference}
Once symbolic representations are constructed, NeSTR uses a large language model to perform inference directly over these structured forms rather than raw text. The symbolic predicates define a constrained and interpretable reasoning space, within which the model applies flexible neural inference. Rather than relying on static fixed-rules, the model learns to match, compare, and compose predicates based on temporal and logical constraints, while drawing on its internal knowledge to identify patterns and support abductive reasoning.

To achieve this, NeSTR introduces an interactive reasoning strategy where symbolic feedback dynamically guides the neural reasoning process. This allows the model to reason step-by-step, leveraging intermediate symbolic signals—such as matched timestamps or transition points—to iteratively refine its inferences. Instead of treating symbolic and neural modules as separate components, this tightly integrated interaction enables controlled and interpretable temporal reasoning.

For example, given the predicates {works\_for(Jaroslav Pelikan, Valparaiso University, 1946-01, 1949-01)} and {works\_for(Jaroslav Pelikan, Concordia Seminary, 1949-01, 1953-01)}, and a question about the employer before \textit{Concordia Seminary}, the model aligns the time intervals and infers that \textit{Valparaiso University} directly preceded \textit{Concordia Seminary} based on matching end and start dates.

More precisely, NeSTR identifies whether a fact $f_i$ is relevant to a query based on interval overlap:

\begin{equation}
f_i \in \mathcal{F}_q \iff t^{(i)}_s \leq t_e^q \quad \text{and} \quad t_s^q \leq t^{(i)}_e
\end{equation}

To further constrain inference to the correct subject, we denote by $s_i$ the subject entity of fact $f_i$, and by $s_q$ the subject entity appearing in the query. A symbolic filtering step ensures subject consistency:
\begin{equation}
\mathcal{F}_q^{(s)} = \{ f_i \in \mathcal{F}_q \mid s_i = s_q \}
\end{equation}

This combination of symbolic filtering and neural pattern recognition supports accurate and generalizable inference over time-dependent information.

\paragraph{Consistency Verification}
Temporal reasoning is prone to errors, and minor temporal misalignments can lead to incorrect conclusions. To mitigate this, NeSTR incorporates a consistency verification step that verifies inferred results against the original symbolic inputs. Within the \verb|<consistency_check>| tags, the language model performs neural inference to evaluate whether the temporal constraints and inferred predicates remain logically coherent.

This step is crucial even in the absence of explicit conflict. It reinforces the validity of reasoning, especially in complex or ambiguous contexts, and enables the model to revise prior inferences if inconsistencies emerge. By systematically verifying intermediate conclusions through neural-symbolic inference, the model ensures logical soundness.

Formally, for each inferred answer $a_i$, the model verifies whether it is logically entailed by at least one context fact:
\begin{equation}
\exists f_j \in \mathcal{F} \quad \text{s.t.} \quad f_j \vdash a_i
\end{equation}
where $\vdash$ denotes a symbolic entailment relation between facts and conclusions.

\paragraph{Abductive Reflection and Correction}

When inconsistencies or missing information are detected, NeSTR invokes an abductive reflection module. In this step, the language model performs neural reasoning over symbolic inputs and prior inference steps to propose minimal and plausible revisions. It generates abductive hypotheses that align with the symbolic structure and temporal context, such as inferring that a date was misinterpreted or that an intermediate event was omitted.

Unlike symbolic systems that halt or fail without resolution in the presence of contradictions, NeSTR uses this neural reflection to actively repair its reasoning and recover from incomplete or ambiguous inputs.

\paragraph{Final Answer Extraction}

After all inference and correction steps are completed, NeSTR generates the final answer based on the verified symbolic reasoning. The answer is written within the \verb|<answer>| tag. Since reasoning and answer generation are explicitly separated, the final output reflects only consistent and validated logic. This separation enhances interpretability and allows users to inspect the complete reasoning trace when needed, which is particularly valuable for questions involving complex temporal constraints.

By structuring temporal reasoning into five modular steps, NeSTR ensures that each stage is transparent, verifiable, and resilient to error. This design enables the system to detect and revise faulty inferences, maintain coherence across temporal facts, and produce answers grounded in symbolic logic. The complete reasoning procedure is detailed in Algorithm~\ref{alg:nestr}.

\begin{algorithm}[tb]
\caption{NeSTR Prompt Strategy}
\label{alg:nestr}
\textbf{Input:} Temporal question $q$, temporal context $c$ \\
\textbf{Output:} Final answer $a$
\begin{algorithmic}[1]
\STATE \textbf{Symbolic Representation}
\STATE Convert facts from context $c$ into symbolic predicates: \texttt{relation(subject, object, start\_time, end\_time)}.
\STATE \textbf{Symbolic Temporal Inference}
\STATE Apply logical rules to infer temporally valid conclusions within \verb|<inference>| tags.
\STATE \textbf{Consistency Verification}
\STATE Validate inferred conclusions against temporal facts within \verb|<consistency_check>| tags.
\STATE \textbf{Abductive Reflection and Correction}
\STATE Revise symbolic representations or inference steps within \verb|<reflection>| tags upon identifying inconsistencies.
\STATE \textbf{Final Answer Extraction}
\STATE Extract and return the concise final answer within \verb|<answer>| tags.
\end{algorithmic}
{\small Note: This is a simplified pseudocode illustration; 
the full prompt is available on GitHub.}
\end{algorithm}

\section{Experiments}

\begin{table*}[t]
\centering
\begin{tabular}{llcccccccccc}
\hline
\textbf{Model} & \textbf{Strategy} & \multicolumn{2}{c}{\textbf{TimeQA-Easy}} & \multicolumn{2}{c}{\textbf{TimeQA-Hard}} & \multicolumn{2}{c}{\textbf{TempReason-L2}} & \multicolumn{2}{c}{\textbf{TempReason-L3}} & \multicolumn{2}{c}{\textbf{Avg}} \\
 &  & EM & F1 & EM & F1 & EM & F1 & EM & F1 & EM & F1 \\
\hline
\multicolumn{12}{l}{\textbf{Literature}} \\
\hline
BigBird$^{\dag}$ & Vanilla & 51.2 & 71.6 & 59.5 & 68.1 & 32.7 & 50.9 & 28.8 & 46.8 & 43.1 & 59.4 \\
FiD$^{\dag}$ & Vanilla & 60.5 & 67.9 & 46.8 & 54.6 & -- & -- & -- & -- & -- & -- \\
T5-large$^{\dag}$ & Vanilla & 63.1 & 71.6 & 59.5 & 68.1 & 32.7 & 50.9 & 28.8 & 46.8 & 46.0 & 59.3 \\
Temp-T5$^{\dag}$ & Vanilla & -- & -- & -- & -- & 31.8 & 49.6 & 26.1 & 43.0 & -- & -- \\
REMEMO-large$^{\dag}$ & Vanilla & 63.7 & 72.3 & 60.5 & 69.3 & 37.4 & 54.9 & 33.4 & 49.3 & 48.8 & 61.5 \\
TG-LLM$^{\dag}$ & CoT & 66.4 & 69.1 & 63.1 & 66.4 & 42.4 & 52.2 & 35.6 & 46.9 & 51.9 & 58.7 \\
ReAct$^{\dag}$ & Few-shot & 45.0 & 55.1 & 28.3 & 34.4 & 39.3 & 45.6 & 42.7 & 50.8 & 38.8 & 46.5 \\
QAaP$^{\dag}$ & Few-shot & 46.3 & 54.4 & 41.7 & 55.3 & 43.7 & 50.1 & 45.3 & 48.3 & 44.3 & 52.0 \\
Event-AL$^{\dag}$ & --  & 63.0 & 73.8 & 61.7 & 70.4 & 55.3 & 62.8 & 58.0 & 59.5 & 59.5 & 66.6 \\
\hline
\multicolumn{12}{l}{\textbf{Open LLMs - Qwen2.5-7B}} \\
\hline
Qwen2.5-7B & Vanilla & 13.0 & 13.0 & 15.1 & 15.1 & 0.03 & 0.03 & 0.07 & 0.06 & 7.1 & 7.1 \\
Qwen2.5-7B & TISER & 86.8 & 92.6 & 64.3 & 71.5 & 61.1 & 69.8 & 72.6 & 77.6 & 71.2 & 77.9 \\
Qwen2.5-7B & NeSTR & 85.1 & 90.2 & 64.8 & 71.2 & 61.5 & 68.6 & 73.1 & 76.7 & 71.1 & 76.7 \\
\hline
\multicolumn{12}{l}{\textbf{Open LLMs - Qwen3-8B}} \\
\hline
Qwen3-8B & Vanilla & 79.8 & 87.9 & 57.2 & 65.3 & 60.0 & 58.5 & 52.2 & 61.0 & 62.3 & 68.2 \\
Qwen3-8B & TISER & 88.8 & 93.4 & 77.1 & 82.5 & 73.7 & 78.4 & 84.3 & 87.5 & 80.9 & 85.4 \\
Qwen3-8B & NeSTR & 89.5 & 94.2 & 77.7 & 83.4 & 79.2 & 83.5 & 84.9 & 87.2 & 82.8 & 87.1 \\
\hline
\multicolumn{12}{l}{\textbf{Open LLMs - Qwen3-14B}} \\
\hline
Qwen3-14B & Vanilla & 78.1 & 87.5 & 61.5 & 72.1 & 45.4 & 59.4 & 58.5 & 67.9 & 60.9 & 71.7 \\
Qwen3-14B & TISER & 90.0 & 94.3 & 82.1 & 87.2 & 75.5 & 80.6 & 81.6 & 85.2 & 82.3 & 86.8 \\
Qwen3-14B & NeSTR & 91.1 & 94.5 & \textbf{82.2} & \textbf{87.3} & 79.5 & 84.6 & \textbf{85.1} & 88.9 & 84.5 & 88.8 \\
\hline
\multicolumn{12}{l}{\textbf{Closed LLMs}} \\
\hline
GPT-4o-mini & Vanilla & 73.4 & 81.7 & 44.7 & 58.5 & 35.7 & 58.2 & 44.8 & 61.0 & 49.7 & 64.9 \\
GPT-4o-mini & TISER & 86.7 & 91.9 & 74.3 & 79.9 & 77.7 & 84.1 & 82.3 & 87.1 & 80.2 & 85.8 \\
GPT-4o-mini & NeSTR & \textbf{93.7} & \textbf{96.4} & {81.7} & {85.9} & \textbf{80.8} & \textbf{86.4} & {84.6} & \textbf{90.0} & \textbf{85.2} & \textbf{89.7} \\
\hline
\end{tabular}
\caption{Exact Match (EM) and token-level F1 scores on in-domain test sets across four temporal reasoning benchmarks. Methods without any special marker are evaluated in a zero-shot setting without fine-tuning. Entries marked with \dag{} are reported from prior work and may involve model training or additional supervision. The ‘Strategy’ column specifies the prompting method used.}
\label{tab:main_results}
\end{table*}

\subsection{Experimental Setup}

\paragraph{Datasets.}
We evaluate NeSTR on four temporal QA benchmarks: \textbf{TimeQA-Easy} and \textbf{TimeQA-Hard}~\cite{chen2021dataset}, and \textbf{TempReason L2} and \textbf{TempReason L3}~\cite{tan2023towards}. TimeQA consists of around 20K questions per subset, covering over 70 temporal relation types; the Easy subset involves direct temporal facts, while the Hard subset requires more complex reasoning. TempReason targets systematic temporal reasoning, with L2 assessing event-time alignment and L3 focusing on event-event comparisons under temporal constraints. Together, the datasets span a wide temporal range and diverse reasoning types, providing comprehensive coverage of temporal QA challenges.

\paragraph{Evaluation Metrics.}

We report {Exact Match (EM)} and token-level {F1} as evaluation metrics following the standard evaluation protocol in prior work ~\cite{bazaga2025learning}. EM checks if the predicted answer matches the reference exactly. F1 measures how much the prediction overlaps with the reference by comparing tokens. These metrics are suitable for temporal reasoning tasks, where correct alignment of events is important.

\paragraph{Baselines.}
We compare NeSTR against both fine-tuned encoder-decoder models and prompt-based frameworks. BigBird~\cite{zaheer2020big} uses sparse attention to scale with long contexts, FiD~\cite{izacard2020leveraging} aggregates segmented evidence via late fusion, T5-large~\cite{raffel2020exploring} is fine-tuned directly for temporal QA, and TempT5~\cite{tan2023towards} incorporates temporal supervision. TG-LLM~\cite{xiong2024large} performs structured reasoning over timelines, REMEMO-large~\cite{yang2023once} retrieves temporally relevant memory, ReAct~\cite{yao2023react} refines answers through iterative search, QAaP~\cite{zhu2023question} reformulates QA as program synthesis, Event-AL~\cite{wu2024event} models temporal and causal links via abduction, Vanilla serves as a simple structured baseline, and TISER~\cite{bazaga2025learning} integrates timeline construction with reflection for improved reasoning.

\paragraph{Implementation Details.}
All models we reproduce are evaluated strictly in a zero-shot setting, without any additional fine-tuning. We use GPT-4o-mini via API as a representative closed-source model. Additionally, we benchmark several open-source models: Qwen2.5-7B (Instruct), Qwen3-8B, and Qwen3-14B. All models are accessed through their respective APIs with the temperature parameter set to 0.1 to ensure deterministic outputs. Each experiment is repeated three times, and the results are averaged to improve reliability. For simplicity and readability, we report average performance over three runs without error bars.

All experiments are conducted in a PyTorch 2.2.0 environment on Ubuntu 22.04, using NVIDIA RTX 4090 for computation. We commit to releasing all source code and evaluation scripts upon publication.

\subsection{Main Results}

Table ~\ref{tab:main_results} summarizes the evaluation results on four temporal reasoning benchmarks, comparing NeSTR with various baselines across multiple models and prompting strategy in a strict zero-shot setting. Our proposed NeSTR consistently achieves superior performance, establishing new state-of-the-art results across these challenging benchmarks.

NeSTR significantly outperforms baseline approaches and alternative prompting methods across different model scales. For instance, when applied to the Closed LLM model, GPT-4o-mini, NeSTR achieves an impressive macro-average F1 score of 89.7, substantially surpassing the vanilla baseline (64.9) and the strong TISER approach (85.8). Notably, NeSTR demonstrates remarkable improvements on the challenging TempReason-L3 dataset, improving F1 from 61.0 (vanilla) to 90.0, underscoring its robust capability for deep, compositional temporal reasoning.

For open-source models such as Qwen3-14B, NeSTR achieves an overall macro-average F1 of 88.8, clearly outperforming the TISER approach (86.4). Specifically, on TempReason-L3, NeSTR reaches an F1 score of 88.9 compared to TISER’s 85.2, indicating its superior temporal abstraction and reasoning capability. Similarly, for Qwen3-8B, NeSTR attains an average F1 of 87.1, substantially exceeding the vanilla (68.2) and TISER (85.4) approaches.

Even on smaller-scale models like Qwen2.5-7B, NeSTR remains highly effective, elevating macro-average F1 from a mere 7.1 under vanilla prompting to a remarkable 76.7, surpassing previous state-of-the-art methods such as REMEMO and Event-AL by significant margins.

These results indicate that NeSTR’s structured neuro-symbolic prompting effectively addresses temporal reasoning limitations inherent in vanilla and alternative prompting strategy. Importantly, NeSTR achieves these gains without any fine-tuning or additional model modifications, highlighting its simplicity, effectiveness, and broad applicability.

Overall, NeSTR not only establishes state-of-the-art performance across multiple challenging temporal benchmarks but also demonstrates excellent scalability and generalization across models of varying sizes and complexities, marking it as a highly effective and practical approach for temporal question answering.

\subsection{Ablation Study}

\begin{table*}[t]
\centering
\begin{tabular}{lcccccccccc}
\hline
\textbf{Setting} & \multicolumn{2}{c}{\textbf{TimeQA-Easy}} & \multicolumn{2}{c}{\textbf{TimeQA-Hard}} & \multicolumn{2}{c}{\textbf{TempReason-L2}} & \multicolumn{2}{c}{\textbf{TempReason-L3}} & \multicolumn{2}{c}{\textbf{Avg. (Macro)}} \\
 & EM & F1 & EM & F1 & EM & F1 & EM & F1 & EM & F1 \\
\hline
Symbolic only     & 88.3 & 93.4 & \textbf{88.2} & \textbf{93.4} & 70.5 & 78.9 & 77.0 & 83.1 & 81.0 & 87.2 \\
w/o Symbol       & 86.5 & 91.4 & 74.2 & 79.1 & 78.2 & 83.8 & 81.6 & 86.5 & 80.1 & 85.2 \\
w/o Consistency Check       & 89.3 & 94.1 & 77.3 & 82.7 & 72.2 & 82.3 & 78.4 & 86.2 & 79.3 & 86.3 \\
w/o Abductive Reflection     & 90.2 & 94.5 & 77.9 & 83.2 & 71.8 & 82.6 & 79.7 & 86.8 & 79.9 & 86.8 \\
\hline
NeSTR(full) & \textbf{93.7} & \textbf{96.4} & {81.7} & {85.9} & \textbf{80.8} & \textbf{86.4} & \textbf{84.6} & \textbf{90.0} & \textbf{85.2} & \textbf{89.7} \\

\hline
Vanilla & 73.4 & 81.7 & 44.7 & 58.5 & 35.7 & 58.2 & 44.8 & 61.0 & 49.7 & 64.9 \\
TISER & 86.7 & 91.9 & 74.3 & 80.0 & 77.8 & 84.1 & 82.3 & 87.1 & 80.3 & 86.8 \\
\hline
\end{tabular}
\caption{Ablation results on four temporal QA benchmarks using {gpt-4o-mini}. We evaluate the contribution of each component in NeSTR by comparing the full model with variants that remove symbolic representation, consistency checking, or abductive reflection, as well as a symbolic-only variant that disables neural reasoning. Results show that each component contributes to overall performance, with the full model achieving the best accuracy. While symbolic-only performs well on simpler datasets, it struggles on complex multi-hop tasks, highlighting the importance of neural components for generalization.}
\label{tab:ablation_study}
\end{table*}

Ablation results in Table~\ref{tab:ablation_study} show the impact of removing individual components of NeSTR’s reasoning framework. We evaluate the performance after systematically disabling (1) the neural component (symbolic-only reasoning), (2) the consistency verification mechanism, and (3) the abductive reflection step. Each component contributes to overall accuracy, confirming the importance of integrated neuro-symbolic reasoning.

The \textbf{Symbolic-only} variant eliminates all neural guidance, relying solely on structured temporal rules and logic-based inference. While competitive on simpler benchmarks like TimeQA-Easy and TimeQA-Hard (achieving 88.2 EM on TimeQA-Hard), its performance decreases notably on complex multi-hop tasks, particularly TempReason-L2 and L3. For instance, EM scores fall from 80.8 to 70.5 on TempReason-L2 and from 84.6 to 77.0 on TempReason-L3, showing that neural components are essential for generalization and deeper compositional reasoning.

The variant \textbf{w/o Symbol} removes explicit symbolic representation and replaces symbolic reasoning with purely natural language-based inference. Compared to the full NeSTR model, this leads to a performance drop, especially on TimeQA-Hard (EM drops from 81.7 to 74.2) and TempReason-L3 (from 84.6 to 81.6). Although still competitive with prompting-based baselines like TISER (80.1 EM vs. 80.3 EM), the decrease confirms that symbolic abstraction plays a key role in enabling structured temporal understanding and precise reasoning over time-sensitive relations. These results indicate that natural language reasoning alone, even with consistency and reflection mechanisms retained, is insufficient for fully modeling the temporal logic required by these tasks.

The variant \textbf{w/o Consistency Check} results in considerable degradation on difficult datasets, especially TimeQA-Hard, where EM decreases from 81.7 to 77.3. This highlights the importance of consistency verification in filtering logically and temporally invalid inferences, helping prevent error propagation and overgeneration.

The variant \textbf{w/o Abductive Reflection} shows moderate but consistent performance loss across TempReason-L2 and L3 datasets. For example, EM drops from 80.8 to 71.8 on L2 and from 84.6 to 79.7 on L3. This suggests that abductive reflection supports reasoning about implicit temporal relationships, improving inference by hypothesizing plausible missing facts.

Overall, the complete NeSTR framework achieves the highest performance across all datasets (85.2 EM / 89.7 F1), confirming that each component contributes complementarily. The symbolic backbone enables core temporal logic, while consistency checks and abductive reflection enhance robustness under uncertainty. These results support NeSTR’s integrated neuro-symbolic approach as key to its strong temporal reasoning capabilities.

\subsection{Effects of Symbolic Representations}


\begin{table}[t]
\centering
\begin{tabular}{lcccc}
\hline
\textbf{Setting} & \multicolumn{2}{c}{\textbf{TimeQA-Hard}} & \multicolumn{2}{c}{\textbf{TempReason-L3}} \\
& EM & F1 & EM & F1 \\
\hline
Vanilla        & 44.7 & 58.5 & 44.8 & 61.0 \\
TISER          & 74.3 & 80.0 & 82.3 & 87.1 \\
\hline
FOL            & 81.6 & \textbf{87.4} & 76.6 & 82.1 \\
Python Dict    & 80.5 & 85.2 & 80.0 & 84.1 \\
\hline
NeSTR   & \textbf{81.7} & {85.9} & \textbf{84.6} & \textbf{90.0} \\
\hline
\end{tabular}
\caption{Comparison of symbolic representation strategies within NeSTR. Both FOL and Python dictionary formats significantly outperform baseline methods, while NeSTR’s full neuro-symbolic integration yields the highest performance.}
\label{tab:different_symbol}
\end{table}

To evaluate the effectiveness of symbolic representations in temporal reasoning, we compare purely natural-language prompting strategies (Vanilla and TISER) with our neuro-symbolic prompting variants, including FOL, Python-style structures, and the full NeSTR pipeline.

As shown in Table~\ref{tab:different_symbol}, both formats significantly outperform baseline methods, underscoring the value of structured abstraction. FOL yields stronger performance on TimeQA-Hard, while the dictionary format excels slightly on TempReason-L3. However, the full NeSTR pipeline, which integrates symbolic structures with neural reasoning, consistently achieves the best results, emphasizing the benefits of neuro-symbolic synergy.

\section{Related Work}

Temporal reasoning remains a major challenge for large language models (LLMs), especially when time information is incomplete, implicit, or changing. Several benchmarks have been developed to study this ability. \textsc{TimeQA} \cite{chen2021dataset, fatemi2024test} first showed that LLMs lag behind humans in answering time-sensitive questions. \textsc{TimeBench} \cite{chu2023timebench} and \textsc{ChronoSense} \cite{islakoglu2025chronosense} further highlight model limitations in understanding temporal intervals, relations, and changing facts, especially in multi-step or compositional reasoning tasks.

To improve temporal understanding, symbolic methods use structured formats to represent time-related facts and apply rule-based reasoning. QAaP \cite{zhu2023question} turns questions and retrieved content into Python-style dictionaries for symbolic checking. TG-LLM \cite{xiong2024large} builds temporal graphs and uses prompting to reason over them. Event-AL \cite{wu2024event} uses event graphs and abductive reasoning to fill in missing links. These methods help with interpretability and temporal alignment, but they often rely on fixed formats and logic, which limits their ability to handle varied language and more flexible reasoning.

Reflective methods offer a more flexible alternative by guiding LLMs to think through intermediate steps. TISER \cite{bazaga2025learning} prompts the model to build and revise timelines, improving consistency. However, without structured symbolic input, it is hard to ensure logical correctness. ABL-ref \cite{hu2025efficient} explores neuro-symbolic reasoning with abductive steps, but focuses on general tasks and has not been used for temporal QA.

Other research aims to improve temporal sensitivity by enriching how time is represented. Timestamp-aware models \cite{dhingra2022time} use time markers to generalize to future events. Temporal knowledge graphs \cite{zhao2025towards,shang2022improving,DBLP:conf/cikm/ZengZ024} and contrastive learning methods \cite{yang2024enhancing} improve event-level reasoning. These models often require extra training or supervision, which limits their use in zero-shot settings.

We build on these ideas with {NeSTR}, a prompting strategy that combines symbolic temporal representation with neural reasoning in a unified design. NeSTR supports multiple symbolic formats (e.g., first-order logic and Python-style structures), and uses the language model to reason over them through abductive reflection and consistency checking. This design allows accurate and interpretable temporal question answering without fine-tuning and performs well across multiple benchmarks.

\section{Conclusion}

In this study, we introduce NeSTR, a neuro-symbolic temporal reasoning prompting strategy designed to enhance large language models' temporal reasoning capabilities through structured symbolic representation, explicit consistency verification, and abductive reasoning. By leveraging symbolic feedback to guide neural inference, NeSTR enables step-by-step reasoning and systematic error correction.

Our experiments show that NeSTR consistently outperforms both prompting-based and fine-tuned baselines across diverse temporal QA benchmarks, especially in complex multi-hop settings. Ablation studies further confirm the contribution of each component, demonstrating that symbolic representation, consistency verification, and abductive correction each play a critical role. We also observe that introducing structured symbolic inputs—regardless of format—consistently leads to significant performance gains over raw language-based reasoning. Finally, attention flow analysis reveals that NeSTR establishes stronger mid-layer connections between answers and questions, enabling more precise and coherent temporal reasoning. In future work, we plan to generalize NeSTR to broader reasoning domains beyond temporal inference.

\section{Acknowledgments}
This work was partially supported by National Natural Science Foundation of China (Nos. U23A20296, 62272469, 62302513), and The Science and Technology Innovation Program of Hunan Province (No. 2023RC1007).


\bibliography{aaai2026}

\end{document}